\documentclass[twoside]{article}

\usepackage[accepted]{aistats2019}
\usepackage[round]{natbib} 

\usepackage[utf8]{inputenc} 
\usepackage[T1]{fontenc}    
\usepackage{hyperref}       
\usepackage{url}            
\usepackage{booktabs}       
\usepackage{amsfonts}       
\usepackage{nicefrac}       
\usepackage{microtype}      
\usepackage{caption}

\usepackage{tikz,amsmath,amssymb}

\usetikzlibrary{calc,bayesnet,shapes.geometric,arrows,chains,matrix,positioning,scopes,calendar,decorations.markings,decorations.pathreplacing,intersections}

\makeatletter
\tikzset{join/.code=\tikzset{after node path={%
      \ifx\tikzchainprevious\pgfutil@empty\else(\tikzchainprevious)%
      edge[every join]#1(\tikzchaincurrent)\fi}}
}
\tikzset{>=stealth',every on chain/.append style={join},
  every join/.style={->}
}

\begin{document}

%

%

\twocolumn[

\aistatstitle{Approximating exponential family models (not single distributions) with a two-network architecture}

\aistatsauthor{Sean R. Bittner \And John P. Cunningham}
\aistatsaddress{ Dept. of Neuroscience \\ Columbia University \And  Dept. of Statistics \\ Columbia University} ]

\begin{abstract}
Recently much attention has been paid to deep generative models, since they have been used to great success for variational inference, generation of complex data types, and more.  In most all of these settings, the goal has been to find a \emph{particular member} of that model family: optimized parameters index a distribution that is close (via a divergence or classification metric) to a target distribution.  Much less attention, however, has been paid to the problem of \emph{learning a model itself}.   Here we introduce a two-network architecture and optimization procedure for learning intractable exponential family models (\emph{not} a single distribution from those models).  These exponential families are learned accurately, allowing operations like posterior inference to be executed directly and generically with an input choice of natural parameters, rather than performing inference via optimization for each particular distribution within that model.
\end{abstract}

\section{Introduction}
Probability models, the fundamental object of Bayesian machine learning, have long challenged researchers with the tradeoff between tractability and expressivity.  Much recent work has focused on deep generative models, which map a latent random variable $w\sim q_0$ through a member of a highly expressive function family $\mathcal{G} = \left\{g_\theta : \theta \in \Theta\right\}$, the composition resulting in an implicit probability model $\mathcal{M} = \left\{ q(g_\theta (w)) : \theta \in \Theta \right\}$.  Choosing $\mathcal{G}$ to be a parameter-indexed family of neural networks has both a rich history \citep{dayan1995helmholtz,mackay1997density}, and has recently been used to produce exciting results for density estimation \citep{uria2013rnade, rippel2013high, papamakarios2017masked}, generation of complex data \citep{Goodfellow:2014aa}, variational inference \citep{Kingma:2013aa, rezende2014stochastic, titsias2014doubly}, and more.  A noted advantage of these deep generative models is that in many cases they make minimal assumptions about the data generative (or posterior inference) process.  

On the other hand, since these models have been chosen to be generic and flexible, they can lack the classic stipulation that a model instantiates existing domain knowledge \citep{gelman2014bayesian, tenenbaum2006theory, mccullagh2002}. 
There are well known drawbacks of fitting such flexible models to finite (albeit large) data sets, which contrast with the bias-variance benefits that come from working in a restricted model space \citep[\S 7.3]{friedman2001elements}.  Work on generalization and compressibility in deep networks suggests that this broad class of function families are indeed quite large, perhaps problematically so \citep{zhou2018compressibility}.  

In this work, we combine the classical wisdom of learning in a restricted model class with modern deep generative approximation methods to learn exponential family models, facilitating the immediate sampling from a member of the model family given an input choice of natural parameters.  This can be particularly powerful when we consider the case of variational inference, where a generative model $p(z)p_\beta(X | z)$ (latent $z$, observed data $X$) is stipulated in the classic sense to embody modeling assumptions (hierarchical model, topic model, Bayesian logistic regression, etc.).  
When performing inference in such a model is intractable, it is increasingly common to deploy an implicit ``recognition network'' model for variational inference \citep{Kingma:2013aa}, which finds a $q_{\theta^*}(z) \in \mathcal{M}$ such that an evidence bound is optimized with respect to the true posterior $p(z|X)$.  
However, it is widely understood that many such true posteriors $p(z|X)$ are exponential families (albeit intractable, due to the choice of sufficient statistics $t(z)$) of the form: $\mathcal{P} = \left\{ \frac{h(z)}{A(\eta)} \exp\left\{ \eta^\top t(z) \right \} : \eta \in H \right\}$ \citep{wainwright2008graphical}.
Some effort has been made to learn single members of exponential families from the mean parameterization \citep{loaiza2017maximum}, but we are focused on learning exponential family models given their natural parameterization.

 Should we be able to learn a tractable approximation to this exponential family model, we would in the very least get the bias-variance benefits of an intelligently restricted model space, and at best would get inference ``for free'' in the sense that we could evaluate approximate posteriors directly without separate optimization for each dataset encountered (a novel form of \emph{amortized inference} \citep{gershman2014amortized,Kingma:2013aa,rezende2014stochastic,stuhlmuller2013learning}).  
 In this paper we aim to learn a restricted model $\mathcal{Q} = \left\{ q(z; \eta): \eta \in H \right\}$ that will be a strict subset of $\mathcal{M}$ and will closely approximate a target exponential family $\mathcal{P}$.  
 Note the critical difference between this aim and much of the literature that seeks to learn a density $q_{\theta}^* \in \mathcal{M}$ (we explore this distinction in depth both algorithmically and empirically).  
 
To proceed, we first specify a set of models $\mathbb{Q} = \left\{ \mathcal{Q}_\phi : \phi \in \Phi \right\}$, from which we can learn a single model $\mathcal{Q}_{\phi^*}$.  We restrict $\Theta$, the parameter space of $\mathcal{M}$, to be itself the image of a second deep \emph{parameter network} family $\mathcal{F} = \left\{f_\phi : \phi \in \Phi\right\}$, such that $\left\{ f_\phi(\eta) : \eta \in H \right\} \subset \Theta$.
Our choice of target $\mathcal{P}$ is an exponential family,  which by definition has \emph{natural} parameterization $\eta \in H$.   Thus, appealingly, we know that $H$ is precisely the correct parameter space for $\mathcal{Q}$ (as it defines $\mathcal{P}$), and that the image of $H$ under $f_\phi$ will be of the correct dimensionality within the codomain $\Theta$. Approximation error between $\mathcal{Q}$ and $\mathcal{P}$ will be caused by the flexibility and learnability of the parameter network $f_\phi$ and the density network $g_{f_{\phi}(\eta)}$.  

We define this two-network architecture, which we term an \emph{exponential family network} (EFN), and we specify a stochastic optimization procedure over a variant of the typical Kullback-Leibler divergence.  We then demonstrate the ability of EFNs to approximately learn exponential families and the benefits of approximating distributions in such restricted model spaces.  Finally we demonstrate the computational savings afforded by this approach when learning the posterior family of point-process latent intensities, given neural spike trains recorded in a neuroscience experiment.

\section{Exponential family networks}

To define exponential family networks (EFNs), we begin with relevant context for our modeling choice of exponential families (\S2.1).  We then describe the network architectural constraint and the background we use to satisfy that constraint (\S2.2). We then introduce EFN in detail, including the optimization algorithm used for learning (\S2.3).  The similarities with variational inference are then explored in depth in (\S2.4).

\subsection{Exponential families as target model $\mathcal{P}$}

We will focus on a fundamental problem setup in probabilistic inference, that of a latent variable $z \in \mathcal{Z}$ with prior belief $p_0(z)$, and where we observe a dataset $X = \left\{x_1,...,x_N\right\} \subset \mathcal{X}$ as conditionally independent draws given $z$.   Updating our belief with data produces the posterior $p(z | X) \propto p_0(z) \prod_{i=1}^N p(x_i | z)$.  This setup is shown as a graphical model in Figure 1A.

If we restrict our attention to priors and likelihoods that belong to exponential families $\mathcal{P} = \left\{ \frac{h(\cdot)}{A(\eta)} \exp\left\{ \eta^\top t(\cdot) \right \} : \eta \in H \right\}$, the posterior can also be viewed as an exponential family, albeit an intractable one \citep{wainwright2008graphical}.  For simplicity we will hereafter suppress the base measure $h(\cdot)$.  Consider:

{\small 
\begin{equation}
 p_0(z) = \frac{1}{A_0(\alpha)} \exp\left\{ \alpha^\top t_0(z) \right\} 
 \label{eq:1}
 \end{equation}
 \begin{equation}
 p(x_i|z) = \frac{1}{A(z)} \exp\left\{ \nu(z)^\top t(x_i) \right \},
\label{eq:2}
 \end{equation} }

where $t(\cdot)$ is the sufficient statistic vector, and $\nu(z)$ is the natural parameter of the likelihood in natural form \citep{robert2007bayesian}.   The posterior then has the form:

\begin{equation}
{\small
  p(z | x_1,...,x_N)  \propto  \exp\left\{ \begin{bmatrix} \alpha \\ \sum_i t(x_i) \\ -N \end{bmatrix}^\top\begin{bmatrix} t_0(z) \\ \nu(z) \\ \log A(z) \end{bmatrix} \right\},
  }
\label{eq:3}
\end{equation}

which again is an intractable exponential family.

To give a concrete example, consider the hierarchical Dirichlet -- a Dirichlet prior $z\sim Dir(\alpha)$ (of dimension $|\mathcal{Z}|$) with conditionally iid Dirichlet draws $x_i | z \sim Dir(\beta z)$, which has been considered historically \citep{mackay1995hierarchical}, and is perhaps most notable for its nonparametric extension \citep{teh2006hdp} (and has relevance for multi-corpus extensions of topic models \citep{blei2003latent, pritchard2000inference}).  
Figure 1B shows the prior for a given $\alpha$ (top), and three examples of datasets that could arise via this generative model (middle).  
A set of basic manipulations shows the hierarchical Dirichlet posterior $p(z|X)$ to be itself an exponential family with natural parameter $\eta = \left[ \alpha -1 , \sum_i \log(x_i) , -N \right]^\top$ and sufficient statistic $t(z) = \left[ \log(z), \beta z , \log(B(\beta z)) \right]^\top$.
The corresponding posteriors are shown in Figure 1B (bottom).  

Note importantly that, because the likelihood was chosen to be an exponential family (which is closed under sampling), this form will not change for any choice of $|Z|$-dimensional hiearchical Dirichlet -- any draw from the prior, any $N$, or any particular realization of observed data $X$ (technically the prior need not be exponential family, but we leave it as such for simplicity).  
The exponential family is clearly sufficient for this property, and the Pitman-Koopman Lemma further clarifies that it is also necessary (under reasonable conditions) \citep[\S3.3.3]{robert2007bayesian}.

The critical observation here is that, if we can approximately learn an intractable exponential family (the model itself), then it becomes trivial to perform posterior inference: we simply use the dataset to index into the natural parameter $\eta$ of the intractable family, and the posterior distribution is produced.  This is the goal of EFNs.

\subsection{Density networks as generic approximating family $\mathcal{M}$}

Deep generative models, which we will use for our approximating model family $\mathcal{M}$, can be defined by any base random variable $w\sim p_0$ mapped through any measurable, parameter-indexed function family  $\mathcal{G} = \left\{g_\theta: \theta \in \Theta\right\}$. We denote the induced density on $z=g_\theta(w)$ as $q_\theta(z)$.   
Though trivial to sample from $q_\theta(z)$ for any choice of family $\mathcal{G}$, we here additionally require that we be able to explicitly calculate $q_\theta(z)$.  
This goal can be readily achieved by designing $\mathcal{G}$ to contain only bijective functions, ideally with a Jacobian form that is convenient to compute. 
Designing that bijective $\mathcal{G}$ as a deep neural network family, as we do here, is a well-established idea that has recently seen many variants and applications \citep{mackay1997density, baird2005one, tabak2010density, rippel2013high, uria2013rnade, rezende2015variational, dinh2016density, papamakarios2017masked, jacobsen2018revnet}.  Specifically, let $z = g_\theta(w) = g_L \circ ... \circ g_1(w)$ for bijective vector-valued functions $g_\ell$ (surpressing $\theta$), and denote $J^\ell_\theta(z)$ as the Jacobian of the function $g_\ell$ at the layer activation corresponding to $z$.  Then we have:

{\small 
\begin{equation}
q_\theta(z) = q_0\left( g_1^{-1} \circ ... \circ g_L^{-1}(z) \right) \prod_{\ell=1}^L \frac{1}{| J^\ell_\theta(z) |}.
\label{eq:4}
\end{equation}}

The specific form of the layers $g_\ell$ can be chosen based on empirical considerations; we clarify our choice in \S3.  For the remainder (and to avoid confusion when we introduce a second network), we call this deep bijective neural architecture the \emph{density network}; this network is shown vertically oriented (flowing from $w$ down to $z$) in Figure 1C.

This density network induces the model $\mathcal{M} = \left\{ q(g_\theta(w)) : \theta \in \Theta \right\}$, which previous work has searched to find a single optimized distribution $q_{\theta^*}$ (such as a posterior or data generative density), on the assumption and subsequent empirical evidence that the target exponential family member is close to (or approximately belongs to) $\mathcal{M}$.   We make the same assumption for the exponential family itself and seek to intelligently restrict $\mathcal{M}$ in order to learn the exponential family.  

\begin{figure}
 {\hspace{-.55cm} %
     \scalebox{0.6}{\input{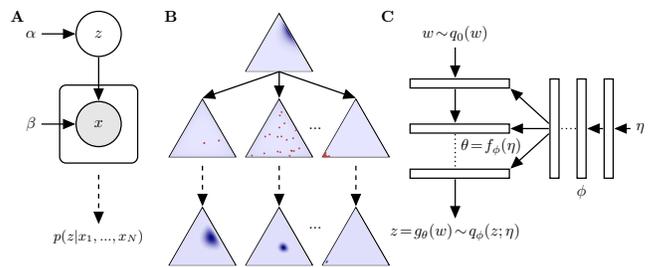}}  
  }

  \caption{(A) Probabilistic graphical model. (B) Hierarchical Dirichlets: a Dirichlet prior with conditionally iid Dirichlet draws.  (top) prior $p_0(z)$, (middle) three sample conditional Dirichlet datasets $X$ of $N=2, N=20, N=100$, and (bottom) three corresponding posteriors that themselves belong to an exponential family $\mathcal{P}$.  (C) Architecture for exponential family network (EFN) -- density network running top to bottom; parameter network running right to left.}
\end{figure}

\subsection{Exponential family networks as approximating model $\mathcal{Q}$}

Having introduced our target model $\mathcal{P}$, an exponential family with natural parameters $\eta \in H$, and the density network family $\mathcal{M}$, we now seek to learn $\mathcal{Q} \approx \mathcal{P}$, where $\mathcal{Q} \subset \mathcal{M}$.  
To do so we will parameterize $\theta$, the parameters of the density network, as the image of a second \emph{parameter network} family $\mathcal{F} = \left\{ f_\phi : H \rightarrow \Theta, \phi \in \Phi\right\}$.   
This network is shown flowing from right to left in Figure 1C.  
Using a second meta-network to aid or restrict network learning has been used in a variety of settings; a few examples include parameterizing the optimization algorithm in the ``learning to learn'' setting \citep{andrychowicz2016learning}, and a more closely related work that used a second network to condition on observations for local latent variational inference \citep{rezende2015variational}, a connection which we explore closely in the following section.

Any choice of parameter network parameters $\phi$ induces a $|H|$-dimensional submanifold (the image $f_\phi(H)$) of the density network parameter space $\Theta$, and as such defines a restricted model $\mathcal{Q}_\phi = \left\{ q_{f_{\phi}}(z; \eta): \eta \in H\right\} \subset \mathcal{M}$; by our choice of $H$ as the natural parameter space of the exponential family target $\mathcal{P}$, this model restriction is at least of the correct dimensionality.
Our goal then is to search over the implied set of models $\mathbb{Q} = \left\{ \mathcal{Q}_\phi : \phi \in \Phi \right\}$ to find an optimal $\phi^*$ such that $\mathcal{Q}_{\phi^*} \approx \mathcal{P}$. 

Given the connections between the exponential family and Shannon entropy, we will measure the error between $\mathcal{Q}_{\phi}$ and $\mathcal{P}$ with Kullback-Leibler divergence.  Consider for the moment a fixed choice of natural parameter $\eta$; we seek to minimize, over $\phi$:

{\small 
\begin{equation} D\left( q_\phi(z;\eta) || p(z;\eta) \right) = \mathbb{E}_{q_\phi} \Bigg( \log q_\phi(z;\eta) - \eta^\top t(z) + \log(A(z)) \Bigg) 
\label{eq:5}
\end{equation}}

which is equivalent to minimizing

{\small 
\begin{multline}  
\mathbb{E}_{q_\phi} \Bigg( \log q_\phi(z;\eta) - \eta^\top t(z) \Bigg) \\ = \mathbb{E}_{q_\phi} \left( \log q_0\left( g_\theta^{-1}(z)\right) + \sum_{\ell=1}^L  \log | J^\ell_\theta(z) | - \eta^\top t(z)) \right),
\label{eq:51}
\end{multline}}

where again we note that $\theta = f_\phi(\eta)$, and thus for a fixed $\eta$, this objective depends only on $\phi$.  Indeed, the target $\eta^\top t(z)$ is linear in $\eta$ (an obvious restatement of the log-linear exponential family form), giving us some hope that we may be able to learn this model.  As a side note, this objective can also produce approximations of the log partition (as the intercept term implied by this linear target), which we have found to be reasonably accurate, though nuanced schemes are likely appropriate \citep{papamakarios2015distilling}.

Of course we seek to approximate not just a single target exponential family member ($p(z;\eta)$ for a fixed $\eta$), but rather the entire model $\mathcal{P} = \left\{p(z;\eta): \eta \in H\right\}$.   For optimization we thus need to introduce a distribution $p(\eta)$ (for stochastic optimization), leading to the objective: 

{\small 
\begin{multline}
\!\arg\!\min_{\!\!\!\!\!\!\!\!\!\!\!\phi} \mathbb{E}_{p(\eta)} \left( D\left( q_\phi(z;\eta) || p(z;\eta) \right)\right) \\ =  \!\arg\!\min_{\!\!\!\!\!\!\!\!\!\!\!\phi}  D\left( q_\phi(z;\eta)p(\eta) || p(z;\eta)p(\eta) \right). 
\label{eq:6}
\end{multline} }

Unbiased estimates of this objective are immediate. $q_\phi(z;\eta)$ is sampled by computing the density network parameters $\theta = f_\phi(\eta)$ (using the parameter network), sampling the latent $w \sim q_0(w)$, and running that $w$ through the density network. $p(\eta)$ is user defined and chosen such that it is trivial to sample.  Stochastic optimization can then be carried out on the estimator:

{\small 
\begin{equation}
\begin{split}
 \mathbb{L}(\phi) = \frac{1}{K}\frac{1}{M}\sum_{k=1}^K & \sum_{m=1}^M \bigg( \log q_0\left( g_{\theta^k}^{-1}\left(z^m\right)\right) \\ & + \sum_{\ell=1}^L  \log | J^\ell_{\theta^k}\left(z^m\right) | - \eta_k^\top t\left(z^m\right) \bigg),
\end{split}
\label{eq:obj}
\end{equation} }

where $\theta^k = f_\phi\left(\eta_k\right)$.  Successful optimization over $\phi$ should thus result in $\mathcal{Q}_{\phi^*} \in \mathcal{M}$ that accurately approximates the target exponential family; that is, $\mathcal{Q} \approx \mathcal{P}$.  We call this two-network architecture and optimization an exponential family network (EFN).   What remains for empirical implementation is to make particular choices of hyperparameters, network layers, and optimization algorithm, which we specify in \S4 below.

\section{Relation to variational inference}

A tremendous amount of work in recent years has gone into variational inference (VI), and its similarity to EFN warrants careful attention. 
In the following, we aim to carefully (and somewhat pedantically) dissect this question.  
As such, though EFN can address any target exponential familiy, to bring us closest to VI let us here restrict the EFN target model $\mathcal{P}$ to be a family of posterior distributions (such as for example the log-Gaussian Poisson example in Section 4.2.)

The typical role of variational inference is to infer an approximate posterior $q_\phi(z) \approx p(z |X)$.  
In this setting, the difference with EFN is stark, in so much as VI learns this single posterior approximation, whereas the main goal of the EFN is to approximate the model $\mathcal{P} = p_\eta(z|X): \eta \in H$: to learn the family of distributions.  
More recently, much focus has gone into the particular instance of VI for local variables $z_i$, for example $\prod_{i=1}^N p(z_i)p(x_i | z_i)$ (such as a variational autoencoder \citep{Kingma:2013aa}) or  $p(u)\prod_{i=1}^N p(z_i|u)p(x_i | z_i)$ (latent Dirichlet allocation being a canonical example \citep{blei2003latent,blei2017variational}), the result of which is often an amortized inference/recognition network that produces a local variational distribution $q_{\phi^*}(z_i | x_i)$.  
This local variational distribution is typically parameterized explicitly: the inference network $\mu_\phi(x_i)$ induces a local parametric distribution, often a Gaussian $q(z_i | x_i) \sim \mathcal{N}\left(z_i; \mu_\phi(x_i)\right)$ \citep[for example]{Kingma:2013aa}.  Viewed this way, local-latent-variable VI methods induce a model $\left\{  q_{\phi^*}(z_i | x_i) : x_i \in X \right\}$ for a finite dataset $X$.   In that sense, EFN and VI are similar `model learning' approaches.
Even more closely, as part of a long-standing desire to add structure to VI beyond mean-field (classically \citep{saul1996exploiting, barber1999tractable}; more recently \citep{hoffman2015stochastic,tran2015copula}, to name but a few), in several cases an inference network has been used to parameterize a deep implicit model (in a two-network inference architecture, to say nothing of whether or not the generative model itself is a deep generative model); closest to the EFN architecture is \citep{rezende2015variational} (cf. Figure 2 of \citep{rezende2015variational} with Figure 1C here).   Thus EFN (when used for posterior families) can be seen as a close generalization of VI.   

Even accepting this VI-as-a-model view, the difference between the finite dataset $X$ and the natural parameter space $H$ persists when viewed at a mechanical level; well-known are the overfitting/generalization issues associated with a finite dataset compared with access to a distribution $p(\eta)$.    Thus one goal of EFN is to allow the model $\mathcal{Q}_{\phi^*} \approx \mathcal{P}$ to be learned in the absence of a finite dataset, such that inference on that dataset can then be executed without concerns of overfitting to that set (and of course without having to run a VI optimization for every new dataset; we will demonstrate this benefit of EFN in the experiments).   Perhaps more importantly, the ``model'' implied by VI is parameterized by $x_i$, and indeed the inference network takes $x_i$ as input. The EFN on the other hand is considerably more general; the posterior includes the natural parameters of the prior (Equation \ref{eq:3}).  This allows the EFN architecture to learn across a more general setting that VI cannot, since any VI inference network is only parameterized by data.  One final difference made clear by Equation \ref{eq:3} is that the observations are given to the EFN \emph{in natural form} (that is, $t(x_i)$, not $x_i$) \citep{robert2007bayesian}.  This choice is a novel insight: by exploiting the known sufficiency of $t(x_i)$ in the target model $\mathcal{P}$, some difference in performance for VI may be observed.  Accordingly, while EFN and VI do at a high level bear multiple similarities, the differences are both material and provoke interesting speculation about means to improve both VI and EFN.

\section{Results}

To investigate the performance of EFNs, we assess approximation fidelity on some tractable exponential families, examine the benefits of learning in a regularized model space, and characterize data analysis scenarios in which training an EFN is computationally advantageous.  First, we test the ability of EFNs to approximate the target model $\mathcal{P}$ when this model is a known, tractable exponential family: this choice provides a simple ground truth and calibrates us to expected performance vs alternatives.   Additionally, tractable exponential families allow us to measure the relative accuracy of single distribution approximations in isolation versus indexed members of trained EFNs.  The main advantage of learning an EFN is to make tractable a previously intractable exponential family (at least approximately).  This confers major benefits in terms of test-time: for example, rather than optimization needing to be run for variational inference with each particular dataset realized from a model class, EFN will allow immediate lookup.  This benefit is orders of magnitude and is not instructive to view, so we show a decision boundary among neural data analysis scenarios, in which training an EFN is computationally advantageous to approximating several distributions through VI optimization individually.  Most often, training an EFN has striking computational advantages. 

To compare model approximations by EFNs to standard methodology, we alternatively train density networks to approximate members of the target model family.  Since $\eta$ will not change, we dispose of the parameter network and train the density network directly over $\theta$ (again with a deterministic choice of a single $\eta$).  When the distribution being approximated is a posterior, this procedure is variational inference.  This is the key comparison for the EFN model, and we refer to this alternative as NF for normalizing flow.

We also must make some particular architectural choices for these experiments.  
We considered a variety of density network architectures.  For each exponential family, we searched through some candidate architectures which consisted of cascades of normalizing flow layers such as planar and radial flows introduced in \citep{rezende2015variational}, a structured spinner flows inspired by \citep{ bojarski2016structured}, and a single affine transformation.
The parameter network was given $\tanh$ nonlinearities. 
In many of the results below we will analyze EFNs across a range of model dimensionality $D$ (that is, $z \in \mathcal{Z} \subseteq \mathbb{R}^D$).    In all cases then we have also $D$ flow layers in the density network (except when the affine transformation is optimal).  In analyses where $D$ was less than 20, 20 flow layers were used.  The number of layers in the parameter network scaled as the square root of $D$, with a minimum of 4 layers, and the number of units per layer scaled linearly from the input to the number of density network parameters. Models were trained using the Adam optimizer algorithm \citep{kingma2014adam}, with learning rates ranging from $10^{-3}$ to $10^{-5}$.  Optimizations ran for at least 50,000 iterations, and completed once there was a subthreshold increase in ELBO. These choices were made so that model performance saturated, and were held constant within comparative analyses.
All code was implemented in tensorflow, and is available at {\tt https://github.com/cunningham-lab/efn}.

\subsection{Tractable exponential families}

Here we study the multivariate Gaussian and Dirichlet families, which offer a known ground truth and intuition about the range of performance that EFN -- learning a model – has with respect to its single-distribution counterpart NF. 
\begin{figure}
  \centering
\includegraphics[width=1.0\linewidth]{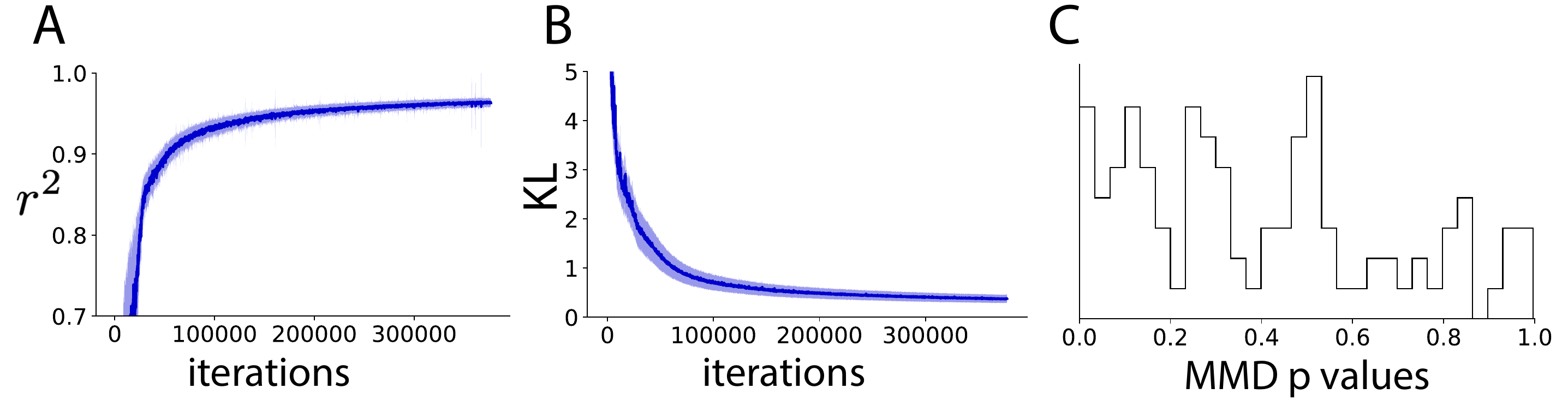}
  \caption{50-dimensional Dirichlet exponential family network.  (A) Distribution of $r^2$ between log density of EFN samples and ground truth across choices of $\eta$ throughout optimization.  (B) Distribution of KL divergence throughout optimization.  (C) Distribution of maximum mean discrepancy p-values between EFN samples and ground truth after optimization.}
\end{figure}
First, to validate the basic EFN approach, we train the $D=50$-dimensional Dirichlet family.  We chose $p(\eta)$, the prior on the $\alpha$ parameter vector of the Dirichlet, as $\alpha_i \sim U\left[.5, 5.0\right]$. The number of $\eta$ samples $K$ at each iteration was $100$, and the minibatch size in $z$ was $M=1000$.   Figure 2 shows a high accuracy fit to this Dirichlet model: Figures 2A and 2B shows rapid convergence to high coefficient of determination $r^2$ and low Kullback-Leibler divergence.  Since we are doing distribution regression, $r^2$ is a convenient metric calculated as the coefficient of determination between the model predictions $\log (q_\phi(z_i; \eta_k))$ and their known targets $\eta_k^\top t(z_i)$.  We can then perform a standard MMD-based kernel two-sample test \citep{gretton2012kernel} between distributions chosen from $\mathcal{P}$ and $\mathcal{Q}_{\phi^*}$: the unstructured distribution of $p$ values clarifies that the EFN model $\mathcal{Q}_{\phi^*}$  is not statistically significantly different than the true target Dirichlet family $\mathcal{P}$ (using a test with 100 samples).

Second, in Figure 3 we consider how this performance scales across dimensionality.  Consider EFN vs NF, where again the only difference is that EFN attempts to learn the entire model (as in $\eta \in H$), whereas NF chooses a single $\eta$ and thus learns a single distribution optimizing the density network parameters $\theta$ directly.  One might expect a noticeable deficit in approximation by EFNs, since they are generalizing the expressivity of the density network across $p(\eta)$.  Accordingly, this deficit is apparent when modeling the multivariate normal family (Fig. 3A).  In low dimensions, we have nearly exact model approximation by EFNs (blue) and distributional approximations by NFs (red).  The distributions learned by NFs were drawn from the same $\eta$ prior as the EFN was trained.  However, as dimensionality increases EFN distributional approximations become significantly worse than the nearly perfect approximations learned by NFs.  The $\eta$ prior of the multivariate normal was specified as an isotropic normal on the mean parameter $\mu_i \sim \mathcal{N}(0, 0.1)$, and an inverse-Wishart distribution on the covariance $\Sigma \sim IW(n, \Psi)$ with degrees of freedom $n=5$ and $\Psi = nD I$.

 \begin{figure}
  \centering
\includegraphics[width=1.0\linewidth]{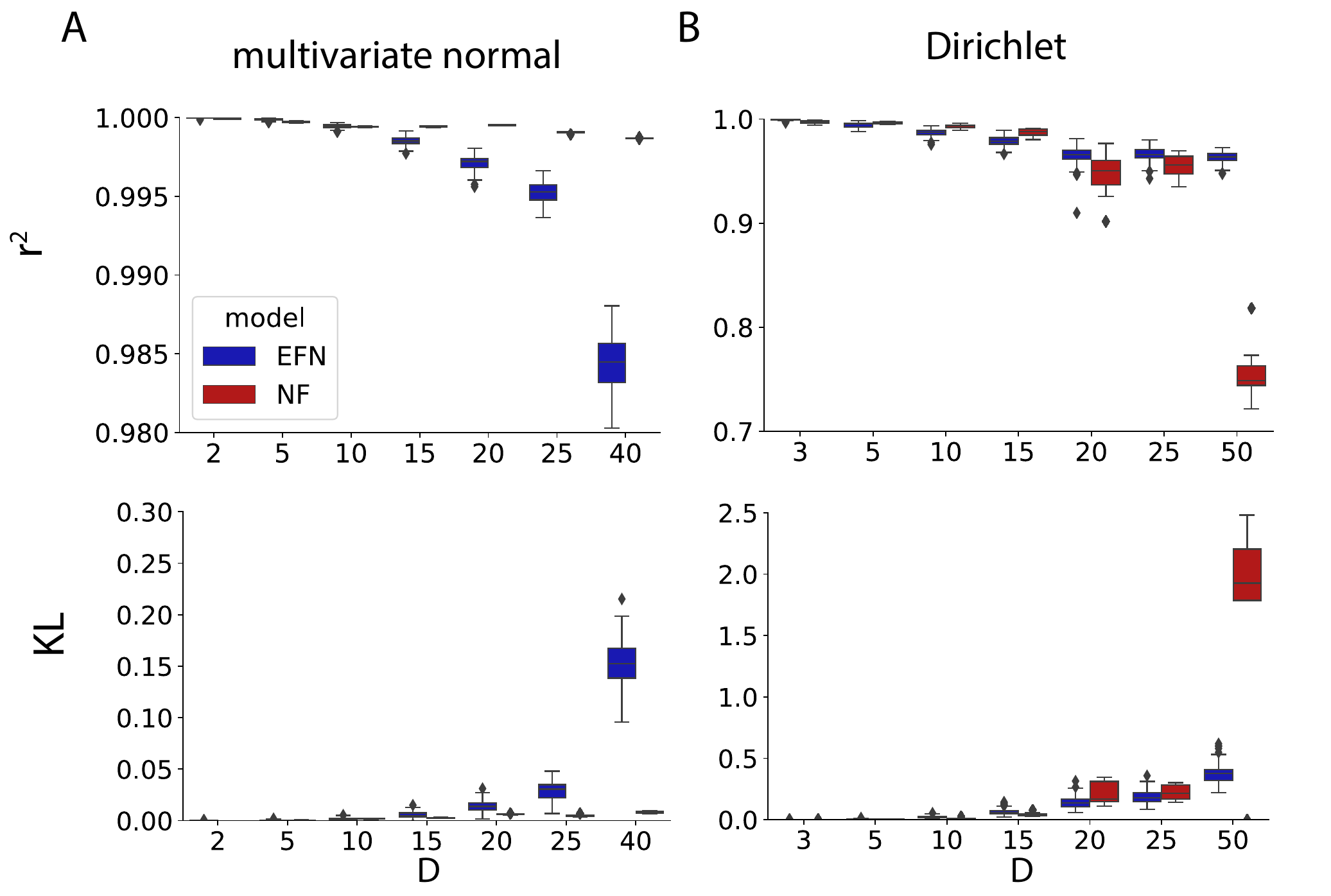}
  \caption{Scaling exponential family networks: $D$ denotes the dimensionality of the family being learned, and comparisons are between EFN and its alternative NF (see text).  (A) Multivariate normal family (B) Dirichlet family.}
\end{figure}

However, learning the model with EFN does not necessarily harm the distributional approximation relative to NF.  In fact, conventional wisdom suggests that learning in a restricted model space is beneficial for regularization.  Here, the expansiveness of the $\eta$ prior determines the necessary degree of generalization of the EFN assigning a weight in the objective to the approximation loss of each distribution.  By requiring the parameter network to learn generalizations of the density network across the $\eta$ prior,  local minima may be avoided that NFs would otherwise be susceptible to.  This is in fact what we see when modeling the Dirichlet distribution (Fig. 3B).  In low dimensions, NF performs better than EFN, but from 20 dimensions and greater, the restricted model space of the EFN confers superior optimization convergence relative to NF, which is more susceptible to local minima.

\subsection{Lookup inference in an intractable exponential family}

Of course the main interest of an EFN is to learn intractable exponential families.  The Gaussian family is the ubiquitous prior for real valued parameters, but it does not match well with the nonnegativity requirements of the intensity measure required of certain distributions, most notably the Poisson.  Log Gaussian Cox Processes have been used numerous times in machine learning, and all have required attention to approximate inference in this fundamentally nonconjugate model. Furthermore, many of these examples have been used to analyze the latent firing intensity of neural spike train data  \citep{cunningham2008fast,cunningham2008inferring,adams2009tractable,gao2016linear}.

\begin{figure}
\centering
\includegraphics[width=1.0\linewidth]{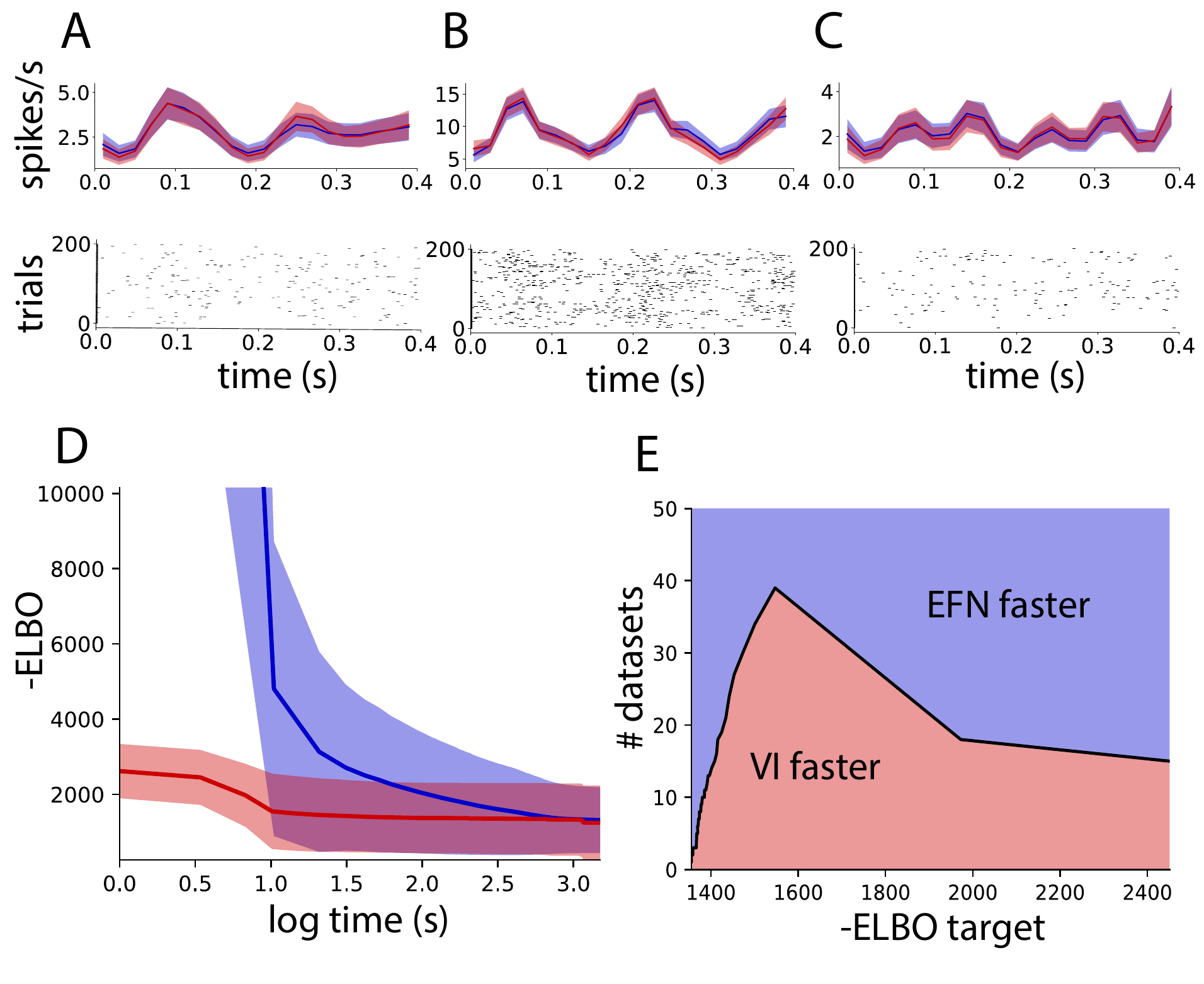}
\captionof{figure}{Lookup inference in a log-Gaussian Poisson model with V1 responses to drift grating stimuli.  (A-C) Top: Inferred latent intensities from a single EFN (blue) or varitional inference (red) run individually for each dataset.  Shading denotes the standard deviation of the posterior.  Bottom: Corresponding V1 spiking responses.  (D) Distribution of -ELBO throughout training across a held out test group of 100 datasets for the EFN (blue), and across 298 datasets fit with NF (red).  (E)  Decision boundary for what number of datasets for a given target approximation accuracy it is advantageous to train an EFN rather than run variational inference individually for each dataset.}
\label{fig:test1}
\end{figure}
We demonstrate the utility of a log-Gaussian Poisson EFN for inferring latent firing intensities of neurons recorded in primary visual cortex of anesthetized macaques in response to 6.25 Hz drift grating stimuli \citep{smith2008spatial}.  200 spike train responses were recorded for each neuron in response to 12 different grating orientations.  Spiking responses were binned into 20ms intervals from 280ms-680ms following stimulus onset (to avoid the effects of transient neural dynamics).  The frequency of the drift grating stimulus suggests a 25ms squared-exponential gaussian process prior on the 20-dimensional latent space, with mean and variance calculated across log firing rates of all neural responses (or "datasets").  Across three experimental subects, 247 neurons with signal-to-noise ratios greater than 1.5 and mean firing rates greater than 1 Hz were considered, resulting in 2,964 total datasets for inferring latent intensities.  By training an EFN on this log-Gaussian Poisson family, we have a model of the posterior distribution for this prior covariance, and some chosen spiking responses.   

We can compare the posterior distribution learned with standard variational inference with NF (red) for a given neuron’s response, to the posterior distribution we get with immediate lookup by supplying the spiking responses of a neuron and the chosen prior (the natural parameters of the posterior) as input to a trained EFN (blue) (Fig. 4A-C).  As a reminder, NF is learning a single member of an exponential family.  If that exponential family is in fact an intractable posterior distribution (such as the hierarchical Dirichlet or log-Gaussian Poisson examples already discussed), then indeed NF is \emph{precisely} performing variational inference with a normalizing flow recognition network, as in \citep{rezende2015variational, dinh2016density, papamakarios2017masked}.  Both EFNs and NFs were trained with 30 planar flow layers.  These posteriors are very similar, and neither appears to fit the data better than the other.  The high quality of these lookup posteriors is an incredible feat by the EFN.  Now that we have trained this EFN, we have immediate posterior inference for all remaining and future neural recordings.

Training an EFN understandably takes more time than an NF (Fig. 4D), but once the EFN is trained we have immediate posterior inference lookup.  If we have a target level of approximation (ELBO target) we can determine when it is faster to get posterior inference on a number of datasets by training an EFN and then using the immediate lookup feature or by running variational inference independently for each distribution.  By computing the amount of computational time it takes to reach the ELBO target on average for both EFN and NF, and then counting how many datasets it would take to learn with NF before eclipsing the training time for the EFN.  This results in a decision boundary (Fig. 4E), where an EFN is more computationally efficient for running posterior inference, and we have infinite computational savings for each additional dataset.  As the ELBO target increases from its minimum value on the right of Fig. 4E, the extra time it takes an EFN to reach this ELBO target relative to NF increases initially.  At some point, the EFN ELBO and NF ELBO distributions begin to converge (Fig. 4D), and the gap of time between learning an EFN and an NF for a given ELBO target begins to decrease.  For some posterior distributions, the EFN learning approach may confer a mean ELBO greater than achievable by traditional variational inference due to the benefits of learning in a restricted model class.  In the case where EFN achieves a greater mean ELBO than NF, it is always advantageous to use EFN.

Our ability to approximate an intractable exponential family model with an EFN is very encouraging.  We have shown in the applied setting of inferring neural firing rates that learning the posterior inference model with an EFN can confer enormous computational savings.  There is nothing unique about this application, insofar as we expect the power of learning exponential family models to translate to applications of intractable exponential family models in other settings.  One can imagine downloading a pre-trained EFN for an intractable exponential family model, and being able to do posterior inference immediately given an arbitrary choice of prior and dataset.

\section{Conclusion}

We have approached the problem of learning an exponential family using a deep generative network, the parameters of which are the image of the natural parameters of the target exponential family under another deep neural network.  We demonstrated high quality empirical performance across a range of dimensionalities, the potential for better approximations when learning in a restricted model space, and computational savings afforded by immediate posterior inference lookup.

\bibliography{BittnerArxiv2019}
\end{document}